\pdfoutput=1
\documentclass{article}
\usepackage{spconf,amsmath,graphicx, amsfonts, bm, multirow}

\title{Deep neural networks on graph signals for brain imaging analysis}
\name{Yiluan Guo, Hossein Nejati, Ngai-Man Cheung}
\address{Singapore University of Technology and Design (SUTD), Singapore 487372}
\begin{document}
\ninept
\maketitle
\begin{abstract}
Brain imaging data such as EEG or MEG are high-dimensional spatiotemporal data 
often degraded by complex, non-Gaussian noise.
For reliable analysis of brain imaging data, it is important to extract discriminative, low-dimensional intrinsic representation of the recorded data.
This work proposes a new method to learn the low-dimensional representations from the noise-degraded measurements.
In particular, our work proposes
a new deep neural network design that integrates graph information such as brain connectivity with fully-connected layers.
Our work leverages efficient graph filter design using Chebyshev polynomial and recent work on convolutional nets on graph-structured data.
Our approach exploits graph structure as the prior side information, localized graph filter for feature extraction and neural networks for high capacity learning.
Experiments on real MEG datasets show that our approach can extract more discriminative representations, leading to improved   accuracy in a supervised classification
task.


\end{abstract}
\section{Introduction}
\label{sec:intro}

\begin{figure}[htb]
\begin{minipage}[b]{.48\linewidth}
  \centering
  \centerline{\includegraphics[width=3.0cm]{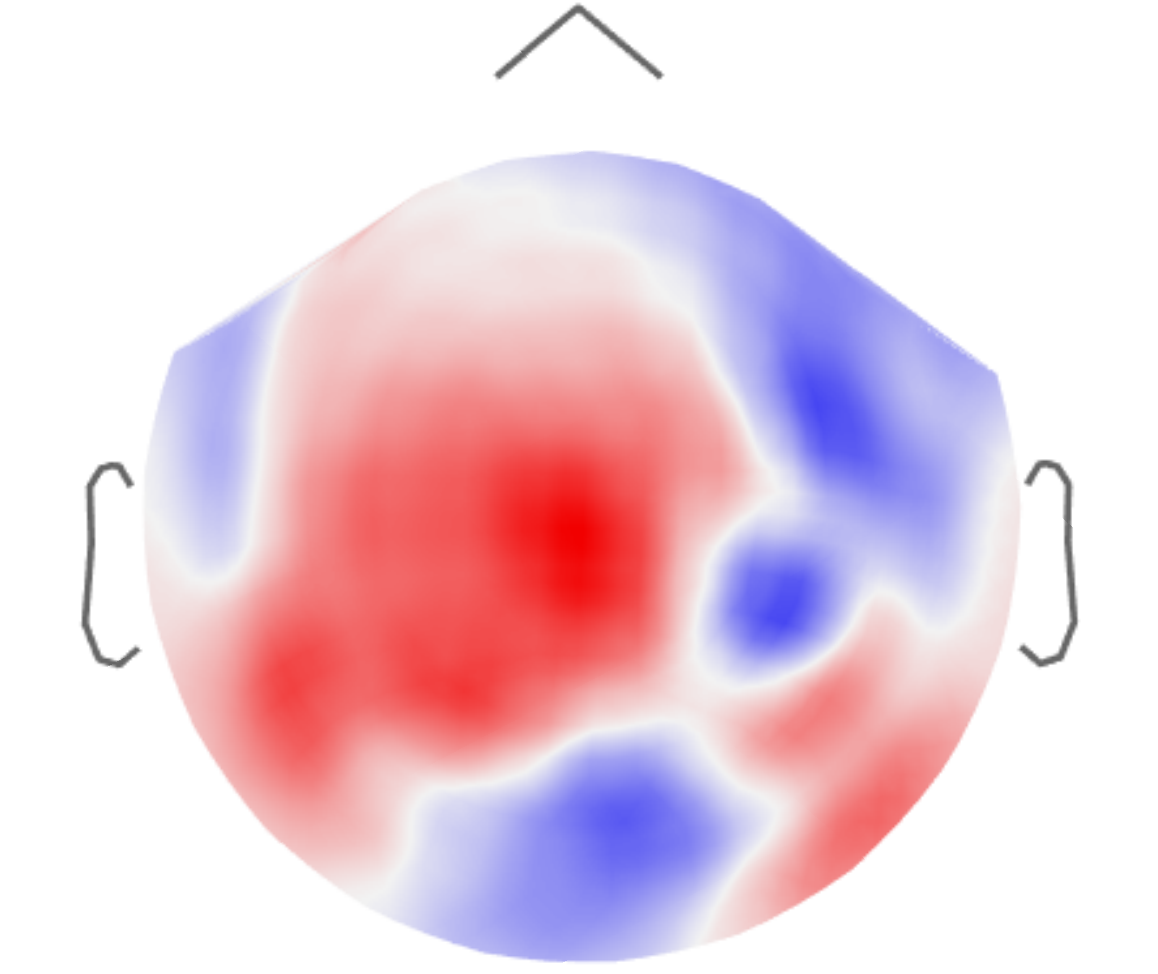}}
  \vspace{0.3cm}
  \centerline{(a) Top view of MEG brain imaging.  }\medskip
\end{minipage}
\hfill
\begin{minipage}[b]{0.48\linewidth}
  \centering
  \centerline{\includegraphics[width=3.0cm]{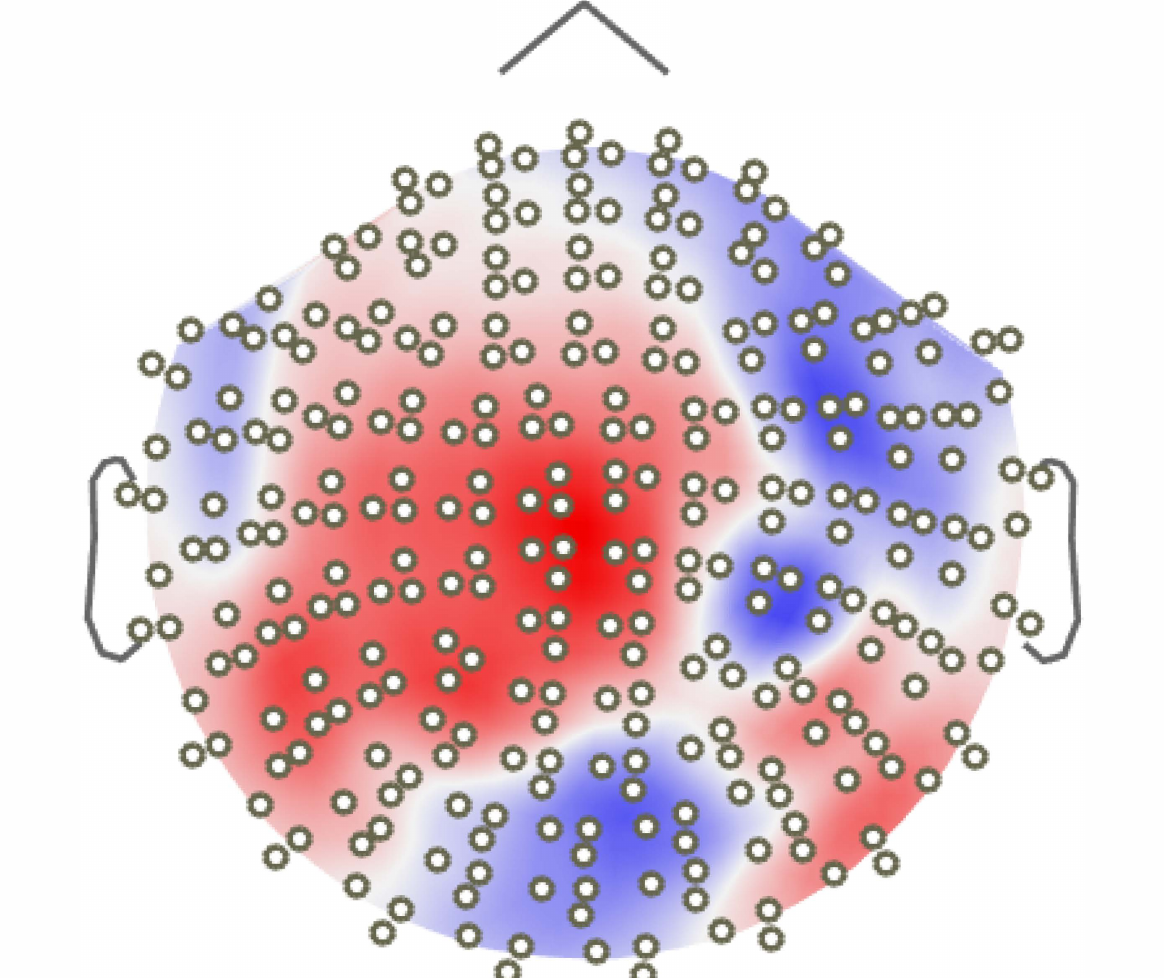}}
  \vspace{0.3cm}
  \centerline{(b) Top view with the sensors. }\medskip
\end{minipage}

\caption{Example of MEG brain imaging data. The color indicates the intensity and directon of the magnetic fields. The nodes in (b) represent the sensors.}
\label{fig1}
\end{figure}

Conventional imaging sensors detect signals lying on regular grids.  On the other hand, recent advances and proliferation in sensing have led to new imaging signals lying on irregular domains.  An example is brain imaging data such as Electroencephalography (EEG) and Magnetoencephalography (MEG).
Some example 
of MEG data used in our experiments is shown in Figure \ref{fig1}(a).
The color in Figure \ref{fig1}(a)
is indicative of the intensity and influx / outflux of magnetic fields. 
The data are different from conventional 2D image data in that they lie irregularly on the brain structure. 
The data are captured by  
a recumbent Elekta MEG scanner with 306
sensors distributed across the scalp to record the cortical activations for 1100 milliseconds (Figure \ref{fig1}(b)). 
Therefore, MEG are high-dimensional spatiotemporal data often degraded by complex, non-Gaussian noise.
For reliable analysis of MEG data, it is important to learn discriminative, low-dimensional intrinsic representation of the recorded data \cite{mwangi:2014,hossein2016}.




Several methods have been applied to perform dimensionality analysis of brain imaging data, e.g.,  principal component analysis (PCA) and its numerous variants (see \cite{mwangi:2014} for a recent review).
In addition,
it has been recognized that there are patterns of anatomical links,  statistical dependencies or causal interactions between distinct units within a nervous system~\cite{bullmore2009complex,hyde2012cross,brovelli2004beta}.  
By modeling brain imaging data as signals residing on {\em brain connectivity graphs},
some methods have been proposed to apply the recent graph signal processing \cite{shuman2013emerging} to analyze brain imaging data \cite{behjat2015anatomically,huang2016graph,rui2016dimensionality,rui:17}.


Deep learning, on the other hand, has achieved breakthroughs in image and video analysis, thanks
to its hierarchical neural network structures with layer-wise non-linear activation and high capacity\cite{lecun2015deep}. As an important deep learning model, autoencoders(AE) / stacked autoencoders(SAE) has achieved state-of-the-art performance in extraction of meaningful low-dimensional representations for input data in an unsupervised way\cite{vincent2010stacked}. However, conventional SAEs fail to take advantage the graph information when the inputs are modeled as graph signals. 

In this work, we propose new AE-like neural networks that tightly integrate graph information for analysis of high-dimensional graph signals such as brain imaging data.
In particular, we propose new AE networks that directly integrate graph models to extract meaningful representations.
Our work leverages efficient graph filter design using Chebyshev polynomial\cite{hammond2011wavelets} and recent work on deep learning on graph-structured data \cite{scarselli2009graph, bruna2013spectral, defferrard2016convolutional, kipf2016semi}.
Among these models, Convolutional Nets(ConvNets) are of great interest since they achieve state-of-the-art 
performance
for images\cite{krizhevsky2012imagenet, he2016deep}
by extracting local features to build hierarchical representations.
Image signals residing on regular grids are suitable for ConvNets. However, the problem to  generalize ConvNets to signals on irregular domains, i.e. graphs, is a challenging one \cite{bruna2013spectral, defferrard2016convolutional, niepert2016learning}. \cite{niepert2016learning} proposed to convert the vertices on a graph into a sequence and extract locally connected regions from graphs, where the convolution is performed in spatial domain. On the contrary, the convolution in \cite{bruna2013spectral} is performed in spectral domain using recent graph signal processing theory \cite{shuman2013emerging}. \cite{defferrard2016convolutional} presented a formulation of ConvNets on graph in spectral domain and proposed fast localized convolutional filters. The filters are polynomial Chebyshev expansions where the polynomial coefficients are the parameters to be learned. \cite{kipf2016semi} applied the first order approximation of \cite{defferrard2016convolutional} and achieved good results on the semi-supervised classification task on social networks.

\begin{figure}[htb]
  \centering
  \centerline{\includegraphics[width=8.5cm]{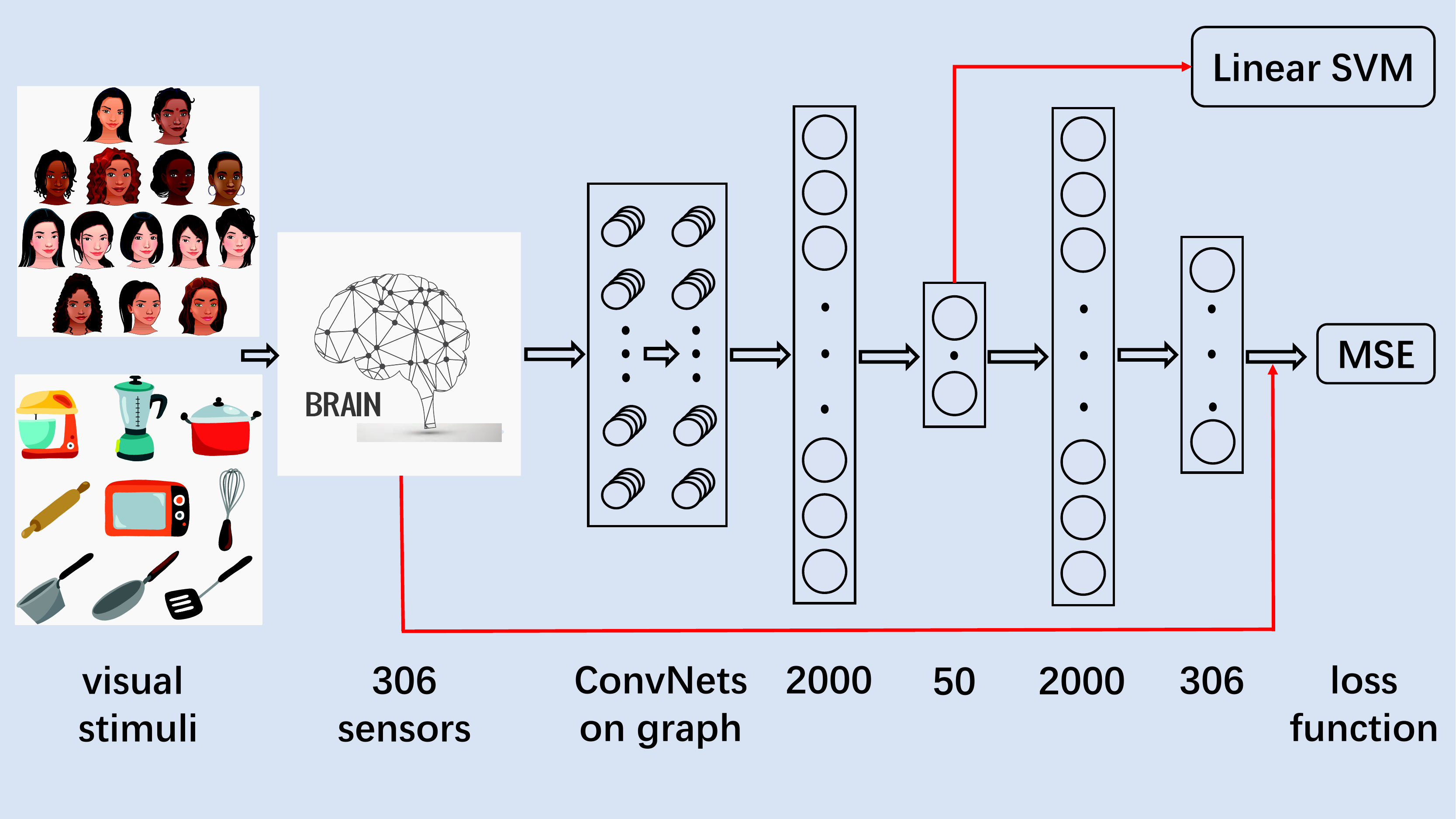}}
\caption{The structure of the proposed method. 
306 MEG sensors are used to record the cortical activations evoked by  
two categories of visual stimuli: face and object.
Recorded high-dimensional MEG measurements and the prior estimated graph are the inputs to the proposed ConvNets on graph. This is followed by an autoencoder with fully connected layers 
of various size.
The entire network is trained end-to-end with mean square error. During testing, 
we extract the activation of the innermost hidden layer and this is subject to a linear SVM to predict whether the subject views face or object.}
\label{fig2}
\end{figure}

This work is inspired by \cite{defferrard2016convolutional, kipf2016semi} but focuses on new AE-like networks to extract meaningful representation in an unsupervised manner.
The proposed method is depicted in Figure \ref{fig2}. First,  brain imaging data is modelled as signals residing on connectivity graphs estimated with causality analysis. Then, the graph signals are processed by the ConvNets on graph, which output high-dimensional, rich feature maps of the graph signals. Subsequently, fully connected layers are used to extract low dimensional representations. During testing, this low-dimensional representations are subject to a linear SVM classifier to evaluate their inclusion of discriminative information. Similar to \cite{kipf2016semi}, we also use the first order approximation in Chebyshev expansions \cite{hammond2011wavelets, defferrard2016convolutional}. However, our network structure is different in that we propose an integration of ConvNets on graph with SAE. The entire network is trained end-to-end in an unsupervised way to learn the low-dimensional representations for the input brain imaging data. In other words, our work is a method of dimensionality reduction. Authors in \cite{Jia2015250} propose to use graph Laplacian to regularize the learning of autoencoder. 
Their work uses a {\em sample graph} to model the underlying data manifold. 
Their approach is significantly different from our work that integrates graph structure into the network. Moreover, it is non-trivial to apply their method to our problem which encodes sensor correlation with a {\em feature graph}.

Our contributions are threefold. First, we model the brain imaging data as graph signals with suitable brain connectivity graphs.
Second, we propose new AE-like network structure that integrates ConvNets on graph with the SAE; the system is trained end-to-end in an unsupervised way. Third, we perform extensive experiments to demonstrate that our model can extract more robust and discriminative representations for brain imaging data. The proposed method can be useful for other high-dimensional graph signals.  

\section{Proposed Method}
\label{sec:proposed method}


We first discuss main results from graph signal processing and ConvNets on graph.  Then we discuss our proposed method.

\subsection{GSP and convolution on graph}
\label{ssec:GSP and GCN}


In conventional ConvNets, local filters are convoluted with signals on regular grids and the filter parameters are learned by back-propagation.
To extend convolution from image / audio signals on regular grids to graph-structured data on irregular domain, recent graph signal processing\cite{shuman2013emerging} provides theoretical results.
In particular,  
we consider an undirected, connected, weighted graph $\mathcal{G} = \{\mathcal{V}, \mathcal{E}, W\}$, which has a number of vertices $|\mathcal{V}| = N$ and an edge set $\mathcal{E}$. $W$ is the symmetric weighted adjacency matrix encoding the edge weights. Graph Laplacian, or combinatorial Laplacian is defined as $L = D - W $, where $D$ is the diagonal degree matrix with diagonal element $D_{ii} = \sum_{j=1}^{N}W_{ij}$. Since $L$ is an symmetric matrix, it can be eigen-decomposed as $L = U\Lambda U^T$ and has a complete set of orthonormal eigenvectors, denoted as $u_l$, 
for $l = 0, 1, ..., N-1$,  and sorted real associated eigenvalues $\lambda_l$, known as the frequencies. In other words, we have $Lu_l = \lambda_lu_l$ for $l = 0, 1, ..., N-1$ and $0 \leq \lambda_0 < \lambda_1 < ... < \lambda_{N-1}$. Normalized graph Laplacian, defined as $\tilde{L} = I - D^{-\frac{1}{2}}LD^{-\frac{1}{2}}$, is also widely used due to the property that all the eigenvalues of it lie in the interval $[0, 2]$. $\{u_l\}$ acts like the Fourier basis in analogy to the eigen-functions of Laplace operator in classical signal processing. The graph Fourier transform(GFT) for a signal $\bm{\mathrm{x}}\in \mathbb{R}^N$ on vertices of the graph
$\mathcal{G}$ is defined as $\tilde{x}(\lambda_l) = \langle u_l, \bm{\mathrm{x}}\rangle = u_l^T\bm{\mathrm{x}}$.  

GFT plays a fundamental role to define filtering and convolution operations for graph signals. Convolution theorem \cite{mallat1999wavelet} states that convolution in spatial domain equals element-wise multiplication in spectral domain. Given 
the signal $\bm{\mathrm{x}}$ and 
a filter $\bm{\mathrm{h}}\in \mathbb{R}^N$ on graph $\mathcal{G}$, the convolution $\ast_{\mathcal{G}}$ between $\bm{\mathrm{x}}$ and $\bm{\mathrm{h}}$ is
\begin{equation}\label{eq1}
\bm{\mathrm{x}}\ast_{\mathcal{G}}\bm{\mathrm{h}} = U((U^T\bm{\mathrm{h}})\odot(U^T\bm{\mathrm{x}})),
\end{equation}
where $\odot$ indicates element-wise multiplication.

In \cite{bruna2013spectral}, the authors proposed spectral neural networks to learn the filters $\bm{\mathrm{h}}$ in spectral domain. 
There are two limitations in this approach. First, it is computationally-intensive to perform GFT and inverse GFT in each feed forward pass. Second, the learned filters using this approach are not explicitly localized, which differ from the filters in conventional ConvNets on images. To overcome these limitations, authors of \cite{defferrard2016convolutional} proposed to use polynomial filters and Chebyshev expansions \cite{hammond2011wavelets}:
\begin{equation}\label{eq2}
\bm{\mathrm{x}}\ast_{\mathcal{G}}\bm{\mathrm{h}} \approx \sum_{k=0}^{K-1} \theta'_kT_k(\hat{L})\bm{\mathrm{x}},
\end{equation}
where $\theta'_k$ are the polynomial filter coefficients to be learned, 
$\hat{L} = \frac{2}{\lambda_{max}}\tilde{L}-I_N$,
and $T_k(\cdot)$ is the Chebyshev polynomial generated recursively. $K$ is the order of the polynomial, which means that the filter is $K$-hop localized. See \cite{hammond2011wavelets,defferrard2016convolutional} for further details.

\subsection{Model structure} 
\label{details about GCN-AE}
Our proposed networks use ConvNets on graph to compute rich features for the input graph signals.  In particular, ConvNets on graph leverage the underlying graph structure of the data to extract local features. Then, we use fully-connected layers and AE-like structure to extract intrinsic representations from the features.

\subsubsection{ConvNets on graph}

\begin{figure}[htb]

  \centering
  \centerline{\includegraphics[width=6.5cm]{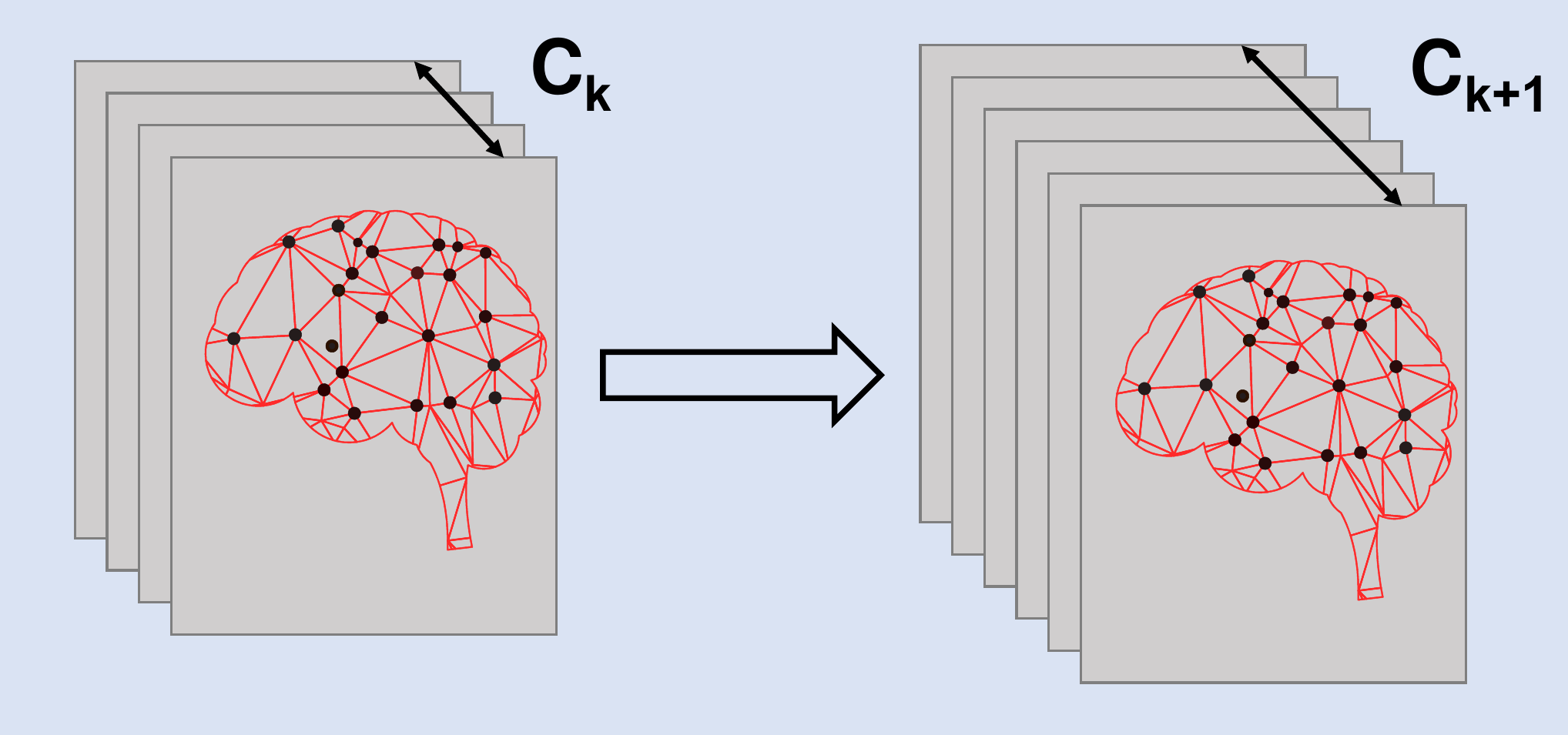}}

\caption{Network structure of the ConvNets on graph. $C_k$ and $C_{k+1}$ are the number of channels at the $k$-th and $(k+1)$-th layers resp.}
\label{fig3}
\end{figure}

The structure of the ConvNets on graph is shown in Figure \ref{fig3}, which integrates the graph information into the neural network. We use the first order approximation of Equation \eqref{eq2} \cite{kipf2016semi}. Since we use normalized Laplacian and all the eigenvalues of it are in the interval [0, 2], we let $\lambda_{max}\approx 2$. Further, we restrict $\theta=\theta'_0 = -\theta'_1$ to  reduce overfitting and computation cost. We also use a renormalization technique proposed in \cite{kipf2016semi}, which converts $I_N + D^{-\frac{1}{2}}AD^{-\frac{1}{2}}$ 
($A$ is the adjacency matrix)
into $\widehat{D}^{-\frac{1}{2}}\widehat{A}\widehat{D}^{-\frac{1}{2}}$, where $\widehat{A} = A +I_N$ and $\widehat{D}$ is the corresponding degree matrix of $\widehat{A}$. 
The reason for renormalization is that the eigenvalues of $I_N + D^{-\frac{1}{2}}AD^{-\frac{1}{2}}$ are in the interval [0, 2], which makes training of this neural network unstable due to gradient explosion\cite{kipf2016semi}. 
After the renormalization, we have\cite{kipf2016semi}
\begin{equation} \label{eq3}
\bm{\mathrm{x}}\ast_{\mathcal{G}}\bm{\mathrm{h}_{\theta}} \approx \theta \widetilde{A} \bm{\mathrm{x}},
\end{equation} 
where $\widetilde{A} = \widehat{D}^{-\frac{1}{2}}\widehat{A}\widehat{D}^{-\frac{1}{2}}$ is the new normalized adjacency matrix for the graph, which takes self-connections into consideration. $\bm{\mathrm{h}_{\theta}}$ indicates the filter $\bm{\mathrm{h}}$ is parameterized by $\theta$, which transforms the graph signal from one channel to another channel.

Recent work \cite{kipf2016semi} uses ConvNets on graph for semi-supervised classification tasks, e.g., semi-supervised document classification in citation networks.
The entire dataset (e.g. full dataset of documents) is modeled as a {\em sample graph} with each vertex representing a sample (e.g., a labeled or unlabeled document).  Therefore, the number of vertices equals to the number of samples.
In their work, they apply two-layer ConvNets on graph to compute a feature vector for each vertex, which is then used to classify a unlabeled vertex. 
In particular, their network processes the whole graph (e.g. entire dataset of documents) as a full-batch.
It is unclear how to scale the design for large dataset.
On the contrary, our network processes individual graph signals in separate passes.
The graph signals are modeled by a {\em feature graph} that encodes the correlation between features. 
The feature graph has $N$ vertices, with $N$ being the dimensionality of a graph signal (for MEG brain imaging data, $N=306$, the number of sensors). Individual low-dimensional representations of the graph signals are subject to classification independently.


In our design, the $k$-th network layer takes as input 
a graph signal $\bm{\mathrm{x_k}} \in \mathbb{R}^{N\times C_k}$, which means that this signal lies on a graph with $N$ vertices and has $C_k$ channels on each vertex.
The output is a graph signal $\bm{\mathrm{x_{k+1}}} \in \mathbb{R}^{N\times C_{k+1}}$. The transformation equation for the $k$-th network layer is 
\begin{equation} \label{eq4}
\bm{\mathrm{x_{k+1}}} = \sigma\bigg(\widetilde{A}\bm{\mathrm{x_k}}\Theta\bigg).
\end{equation}
Here $\sigma (\cdot)$ is the element-wise non-linear activation function; $\Theta \in \mathbb{R}^{C_k\times C_{k+1}}$ is 
the parameter matrix to be learned.  Note that $\Theta$ generalizes the $\theta$ in 
(\ref{eq3}) for multiple channels.   $\Theta$ has dimension $C_k\times C_{k+1}$: the input signal with $C_k$ channels is transformed into one with $C_{k+1}$ channels. 
With the normalized adjacency matrix 
$\widetilde{A}$ in (\ref{eq4}), the network layer considers correlation between individual vertices and their 1-hop neighbors.
To take $m$-hop neighbours into account, $m$ layers need to be stacked. In our experiment, we only stack two ConvNets on graph layers and this shows competitive performance. 
Note that
$\widetilde{A}$ plays the role of specifying the receptive field for one feature: one feature is convoluted with its neighbours on the graph with different weights, which are determined by the nonzero value of $\widetilde{A}$. This is different from conventional ConvNets for images, where the weights is learned by back-propagation. In our work, the neural networks instead learn the weights for transforming the channels of the input graph signal. Note that with the non-linear activation function, the transformation in each network layer is not simply multiplication.


In comparison, conventional neural networks can also expand or compress number of the channels with $1\times 1$ convolution. Specifically, this is the ConvNets on graph when $\widetilde{A} = I$, where $I$ is the identity matrix. This is a limited model due to small kernel size. In fact, when $\widetilde{A}=I$, the ConvNets on graph reduce to fully connected layers in a conventional AE. Similarly, removing the non-linearity activation function limits the model capacity.
Even with larger receptive field for one feature, the output becomes linear combination of the neighbours on graph of this feature. We observe in our experiment (Section \ref{sec:Experiment}) that without $\widetilde{A}$ and non-linearity activation function, our design has similar performance as conventional AEs.

\subsubsection{Fully connected layers and loss function}

After $k$ layers of ConvNets on graph, we obtain a graph signal $\bm{\mathrm{x_k}} \in \mathbb{R}^{N\times C_k}$ of features.
Each row vector is the multichannel feature of one vertex. We concatenate the row vectors and obtain $\bm{\mathrm{x_k}} \in \mathbb{R}^{N\cdot C_k} $ as the output of ConvNets on graph. Since our goal is to extract low dimensional and semantically discriminative representations for each signal in an unsupervised way, we introduce stacked autoencoder(SAE) \cite{vincent2010stacked} here. SAE has been shown by recent research that it consistently produces high-quality semantic representations on several real-world datasets\cite{le2013building}. The difference between our work and SAE is that SAE takes the original signal as input while our work takes as input the high dimensional, rich feature map of the graph signal, which is the output of ConvNets on graph. The dimension of the SAE output $\bm{\mathrm{y}}$ is the same as the original signal. The training of the entire network  is 
end-to-end
by minimizing mean square error between input $\bm{\mathrm{x}}$ and $\bm{\mathrm{y}}$, i.e. $\lVert \bm{\mathrm{x}} - \bm{\mathrm{y}} \rVert^2_2$.

\section{Experiment}
\label{sec:Experiment}

\subsection{Datasets}
We test our model on real MEG signal datasets. The MEG signals record the brain responses to two categories of visual stimulus: human face and object. The subjects were shown 322 human-face and 197 object images randomly while MEG signals were collected by 306 sensors on the brain. The signals were recorded 100ms before the stimulus and until 1000ms after the stimulus onset. Each image was shown to the subjects for 300ms. We focus on MEG data from 96ms to 110ms after the visual stimulus onset, as it has
been recognized that the cortical activities in this duration contain rich information \cite{thorpe:1996}. 
We model the MEG signals as graph signals by regarding the 306 sensor measurements as signals on a graph of 306 vertices. The underlying graph, which represents the complex brain network\cite{guye2010graph}, is estimated 
by Granger Causality connectivity(GCC) analysis using the Matlab open-source toolbox BrainStorm\cite{tadel2011brainstorm}. Note that we have to renormalize the connectivity matrix following our discussion in Section \ref{details about GCN-AE}.

\subsection{Implementation}
We use TensorFlow\cite{abadi2016tensorflow} to implement our networks. The numbers of channels for the two-layer ConvNets on graph are set to be 16 and 5. The subsequent fully-connected layers have dimension $d-2000-50-2000-306$, where $d$ is the dimension  after concatenation of the row vectors of the output of ConvNets. Adam\cite{kingma2014adam} is adopted to minimize the MSE with learning rate 0.001. Dropout\cite{srivastava2014dropout} is used to avoid overfitting. We also include the $L2$ regularization in the loss function for the fully connected layers. For comparison, we train two different SAEs with the same schemes. After training all the networks for 300 epochs, we use linear SVM to predict whether the subject viewed face or object based on the 50-dimensional representation of the original MEG imaging data. We use 10-fold cross validation and report the average accuracy. All the experiments are performed on each subject separately. 

\subsection{Results}

We compare our results with several unsupervised dimensionality reduction methods: PCA, GBF, Robust PCA and SAE. PCA is a commonly used dimensionality reduction technique by projecting data to the axis with first $n$ largest variance. GBF \cite{egilmez2014spectral,rui2016dimensionality}  projects the MEG signals to a linear subspace spanned by the first $n$ eigenvectors of the normalized graph Laplacian. Robust PCA(RPCA) \cite{candes2011robust} decomposes the data into two parts: low rank representation and sparse perturbation.
For non-linear transformation, we test two SAEs, one is with symmetric structure $306-2000-50$ and the other $306-5000-1500-2000-50$.

\begin{table}[htb]
\setlength{\abovecaptionskip}{0cm}
\setlength{\belowcaptionskip}{-10pt}
\centering
\caption{Average classification accuracy with different methods on MEG brain imaging data.}
\label{tab1}
\begin{tabular}{|c|c|c|c|}
\hline
\multirow{2}{*}{Method} & \multicolumn{3}{c|}{Accuracy}                       \\ \cline{2-4} 
                        & subject A       & subject B       & subject C       \\ \hline
original data           & 0.6482          & 0.6015          & 0.6338          \\ \hline
PCA                     & 0.6529          & 0.5957          & 0.6100          \\ \hline
RPCA                    & 0.6656          & 0.5925          & 0.6186          \\ \hline
GBF                     & 0.6638          & 0.6026          & 0.5970          \\ \hline
2-layer AE              & 0.6610          & 0.5983          & 0.6302          \\ \hline
4-layer AE              & 0.6693          & 0.5939          & 0.6323          \\ \hline
proposed model          & \textbf{0.6833} & \textbf{0.6414} & \textbf{0.6435} \\ \hline
\end{tabular}
\end{table}
 
The results are shown in Table \ref{tab1}. It can be observed that accuracy for the original 306-dimensional data is inferior or similar to other methods.  Thus, it is advantageous to perform dimensionality reduction and feature extraction. Improvement using PCA is limited as it  is not robust to the existing non-Gaussian noise. 
For subject A and B, RPCA achieves similar result as GBF, which leverages Granger Causality connectivity(GCC) of subjects' brain as side information. PCA, RPCA and GBF are linear transformations failing to capture the non-linearity property of the brain imaging data, which limits the performance. SAEs with 2 layers and 4 layers also outperform PCA by introducing non-linear transformation. \cite{he2016deep} has shown that increasing the depth of networks helps improve performance by a large margin. Nevertheless, the results are similar for the two SAEs. We conjecture that the optimization stops at saddle points or local minima\cite{dauphin2014identifying}.  Our proposed model achieves the highest accuracy comparing to other methods. The reasons are that our approach 1) considers connectivity as the prior side information and 2) uses neural networks with high capacity to learn the discriminative representation. 


\subsection{Discussion}

\subsubsection{Contribution of the graph}
We may ask whether the graph information is truly helpful and necessary for this task. To answer this question and better understand the importance and necessity of incorporating the graph information in the neural networks, we replace the graph adjacency matrix estimated by GCC with an identity matrix and a random symmetric matrix and train the model.  Table \ref{tab2} shows that GCC indeed helps the networks to extract expressive features. Replacing GCC with identity matrix ignores  the prior feature correlation, resulting in accuracy similar to SAEs. Random symmetric matrix  confuses the neural networks and thus the accuracy drops drastically. 

\begin{table}[htb]
\centering
\caption{Classification accuracy with different adjacency matrix.}
\label{tab2}
\begin{tabular}{|c|c|c|c|}
\hline
\multirow{2}{*}{Graph} & \multicolumn{3}{c|}{Accuracy}     \\ \cline{2-4} 
                       & subject A & subject B & subject C \\ \hline
GCC                    & 0.6833    & 0.6414    & 0.6435    \\ \hline
Identity Matrix        & 0.6616    & 0.6052    & 0.6213    \\ \hline
Random Matrix          & 0.5941    & 0.5589    & 0.5332    \\ \hline
\end{tabular}
\end{table}
\subsubsection{Contribution of nonlinear transformation}
Since we expand our single channel MEG data to multiple channels, there is concern that the transformation is a trivial multiplication with a scaler in graph ConvNets. Therefore, in this experiment, we remove the non-linearity activation function in ConvNets on graph. By doing this, the outputs of the graph ConvNets become the average of the input weighted by the graph adjacency matrix, which is equivalent to linear combination of the inputs. Thus, the accuracy should be similar to SAEs.  This can be observed in Table \ref{tab3}. With non-linear activation function, ConvNets on graph can fully exploit the graph information. 

\begin{table}[htb]
\centering
\caption{Classification accuracy with different activation function.}
\label{tab3}
\begin{tabular}{|c|c|c|c|}
\hline
\multirow{2}{*}{Activation Function} & \multicolumn{3}{c|}{Accuracy}     \\ \cline{2-4} 
                                     & subject A & subject B & subject C \\ \hline
Non-linear                            & 0.6833    & 0.6414    & 0.6435    \\ \hline
Linear                               & 0.6656    & 0.6016    & 0.6132    \\ \hline
\end{tabular}
\end{table}
  
\section{Conclusion}
\label{conclusion}
In this work, we  propose  AE-like deep neural network that integrates ConvNets on graph with fully-connected layers.  The proposed network is used to learn the low-dimensional, discriminative representations for brain imaging data.  Experiments on real MEG datasets suggest that our design extracts more discriminative information than other advanced methods such as RPCA and autoencoders. The improvement is due to the exploitation of graph structure as side information. For future work, we apply recent graph learning techniques \cite{DiscGlassoJYK,SparsityDictionaryHPM} to improve the estimation of the underlying connectivity graph.  
Moreover, we  address the problem of deploying the networks for real-time analysis in brain computer interface applications.
Furthermore, we explore applications of our ConvNets on graph integrated AE for other image / video applications \cite{cheung:07,fang:14}.

\bibliographystyle{IEEEbib}
{
\footnotesize
\bibliography{refs}
}
\end{document}